\def\year{2020}\relax
\documentclass[letterpaper]{article} % DO NOT CHANGE THIS
\usepackage{aaai20}  % DO NOT CHANGE THIS
\usepackage{times}  % DO NOT CHANGE THIS
\usepackage{helvet} % DO NOT CHANGE THIS
\usepackage{courier}  % DO NOT CHANGE THIS
\usepackage[hyphens]{url}  % DO NOT CHANGE THIS
\usepackage{graphicx} % DO NOT CHANGE THIS
\usepackage{graphicx}  % DO NOT CHANGE THIS
\usepackage{amsmath}
\usepackage{amssymb}
\usepackage{array}
\usepackage[ruled,vlined]{algorithm2e}

\DeclareMathOperator*{\argmin}{arg\,min}

\title{On the Anatomy of MCMC-Based Maximum Likelihood Learning of Energy-Based Models}
\author{Erik Nijkamp*, Mitch Hill*, Tian Han, Song-Chun Zhu, and Ying Nian Wu\\
UCLA Department of Statistics\\
8117 Math Sciences Bldg. \\
Los Angeles, CA 90095-1554\\
(*\emph{equal contributions})
}

\begin{document}
\maketitle
\begin{abstract}
This study investigates the effects of Markov chain Monte Carlo (MCMC) sampling in unsupervised Maximum Likelihood (ML) learning. Our attention is restricted to the family of unnormalized probability densities for which the negative log density (or energy function) is a ConvNet. We find that many of the techniques used to stabilize training in previous studies are not necessary. ML learning with a ConvNet potential requires only a few hyper-parameters and no regularization. Using this minimal framework, we identify a variety of ML learning outcomes that depend solely on the implementation of MCMC sampling.

On one hand, we show that it is easy to train an energy-based model which can sample realistic images with short-run Langevin. ML can be effective and stable even when MCMC samples have much higher energy than true steady-state samples throughout training. Based on this insight, we introduce an ML method with purely noise-initialized MCMC, high-quality short-run synthesis, and the same budget as ML with informative MCMC initialization such as CD or PCD. Unlike previous models, our energy model can obtain realistic high-diversity samples from a noise signal after training. 

On the other hand, ConvNet potentials learned with non-convergent MCMC do not have a valid steady-state and cannot be considered approximate unnormalized densities of the training data because long-run MCMC samples differ greatly from observed images. We show that it is much harder to train a ConvNet potential to learn a steady-state over realistic images. To our knowledge, long-run MCMC samples of all previous models lose the realism of short-run samples. With correct tuning of Langevin noise, we train the first ConvNet potentials for which long-run and steady-state MCMC samples are realistic images.
\end{abstract}

\section{Introduction}

\subsection{Diagnosing Energy-Based Models}\label{sec:prior}

\begin{figure}[ht!]
	\centering
	\includegraphics[width=.45\textwidth]{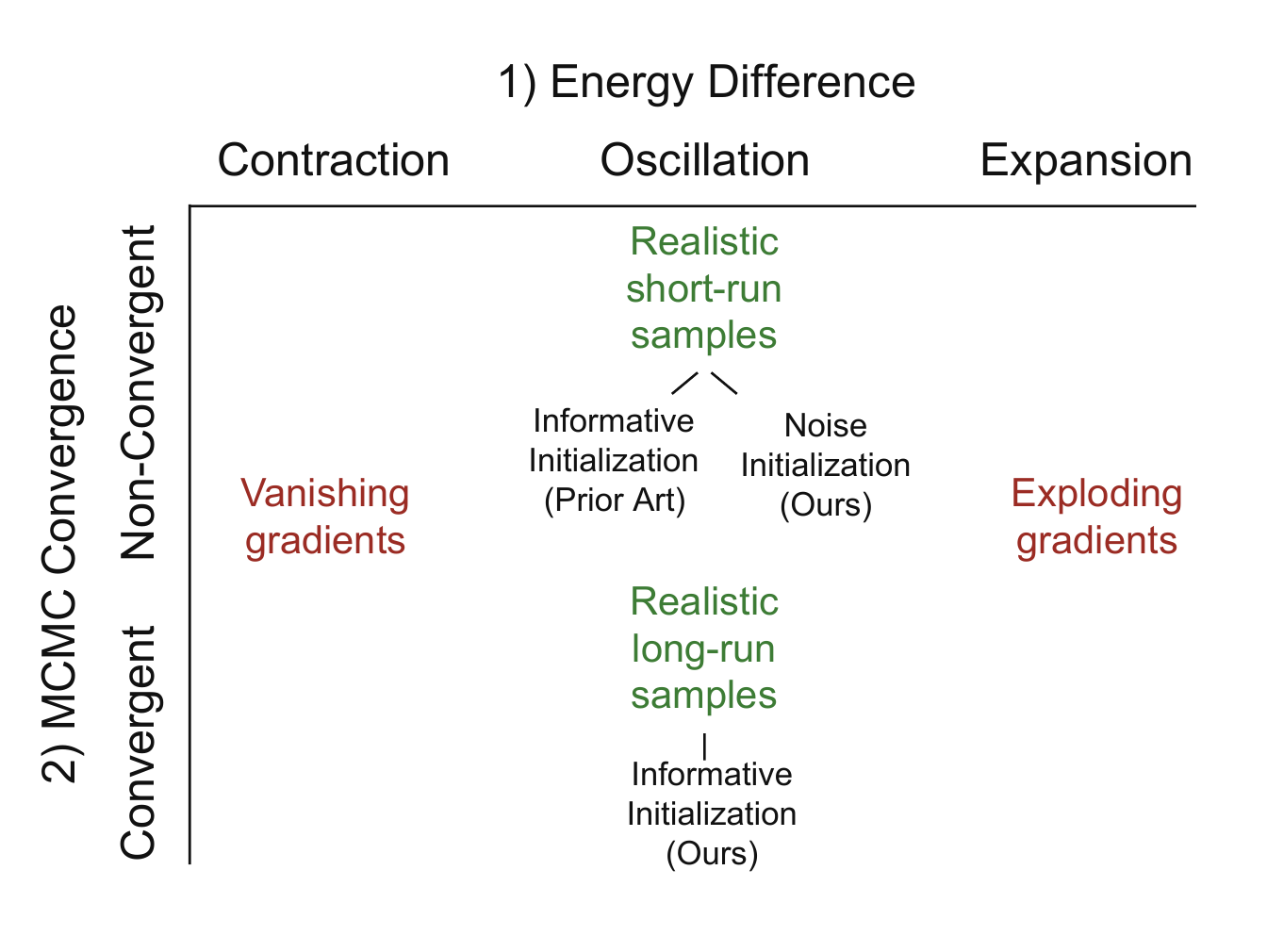}
	\caption{Two axes characterize ML learning of ConvNet potential energy functions: 1) energy difference between data samples and synthesized samples, and 2) MCMC convergence towards steady-state. Learning a sampler with realistic short-run MCMC synthesis is surprisingly simple whereas learning an energy with realistic long-run samples requires proper MCMC implementation. We propose: a) short-run training with noise initialization of the Markov chains, and b) an explanation and implementation of correct tuning for training models with realistic long-run samples.}
	\label{fig:idea}
\end{figure}

Statistical modeling of high-dimensional signals is a challenging task encountered in many academic disciplines and practical applications. We study image signals in this work. When images come without annotations or labels, the effective tools of deep supervised learning cannot be applied and unsupervised techniques must be used instead. This work focuses on the unsupervised paradigm of the energy-based model (\ref{eqn:gibbs_density}) with a ConvNet potential function (\ref{eqn:deepframe_energy}).

Previous works studying Maximum Likelihood (ML) training of ConvNet potentials, such as \cite{xie2016theory,xie2016cooperative,gao2018learning}, use Langevin MCMC samples to approximate the gradient of the unknown and intractable log partition function during learning. The authors universally find that after enough model updates, MCMC samples generated by short-run Langevin from \emph{informative initialization} (see Section~\ref{subsec:mcmc_priming}) are realistic images that resemble the data.

\begin{figure}
	\centering
	{\small1) Non-Convergent ML} \\
	\hspace{-.6cm}\includegraphics[width=.37\textwidth]{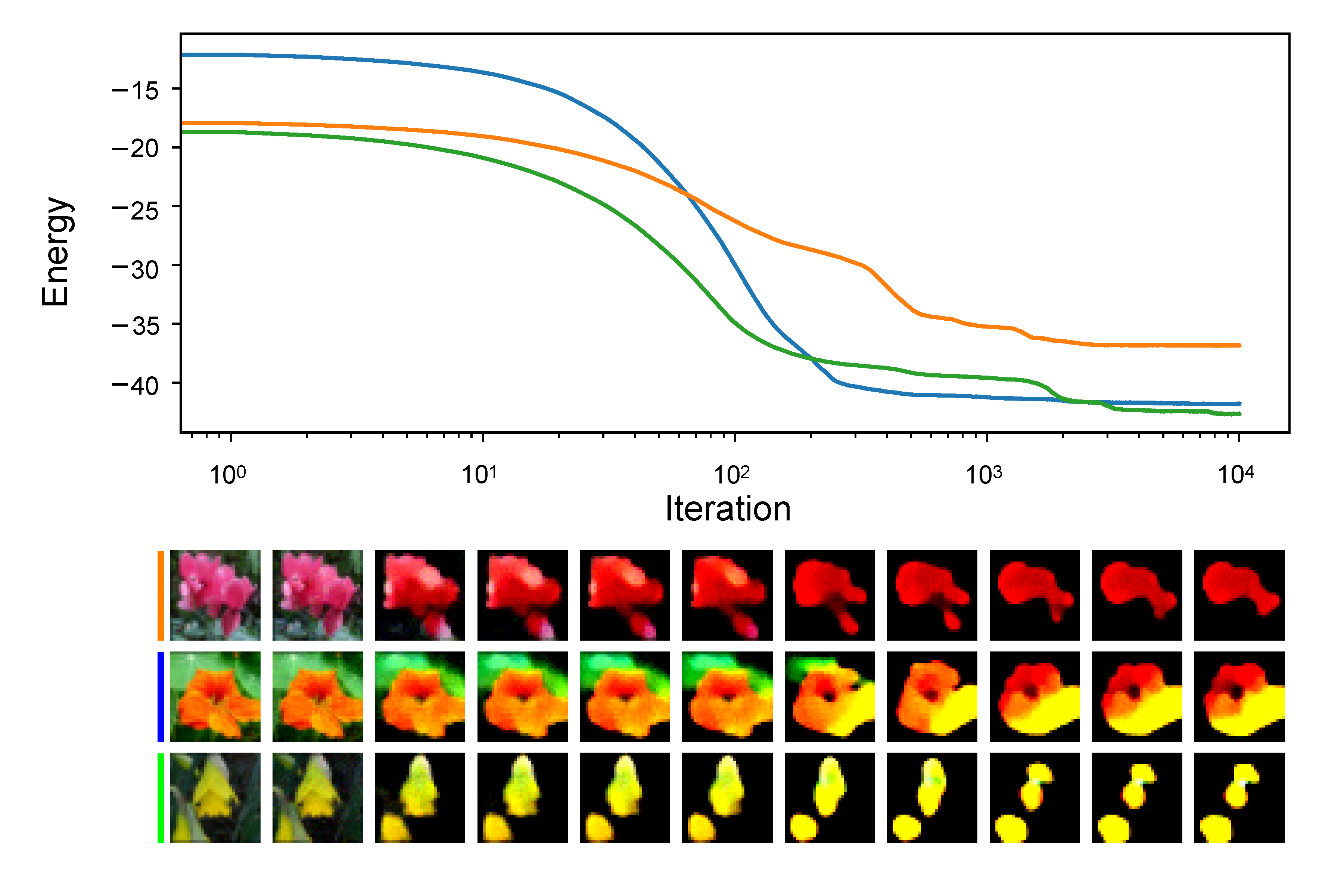} \\
	
	\medskip 
	
	{\small2) Convergent ML} \\
	\hspace{-.6cm}\includegraphics[width=.37\textwidth]{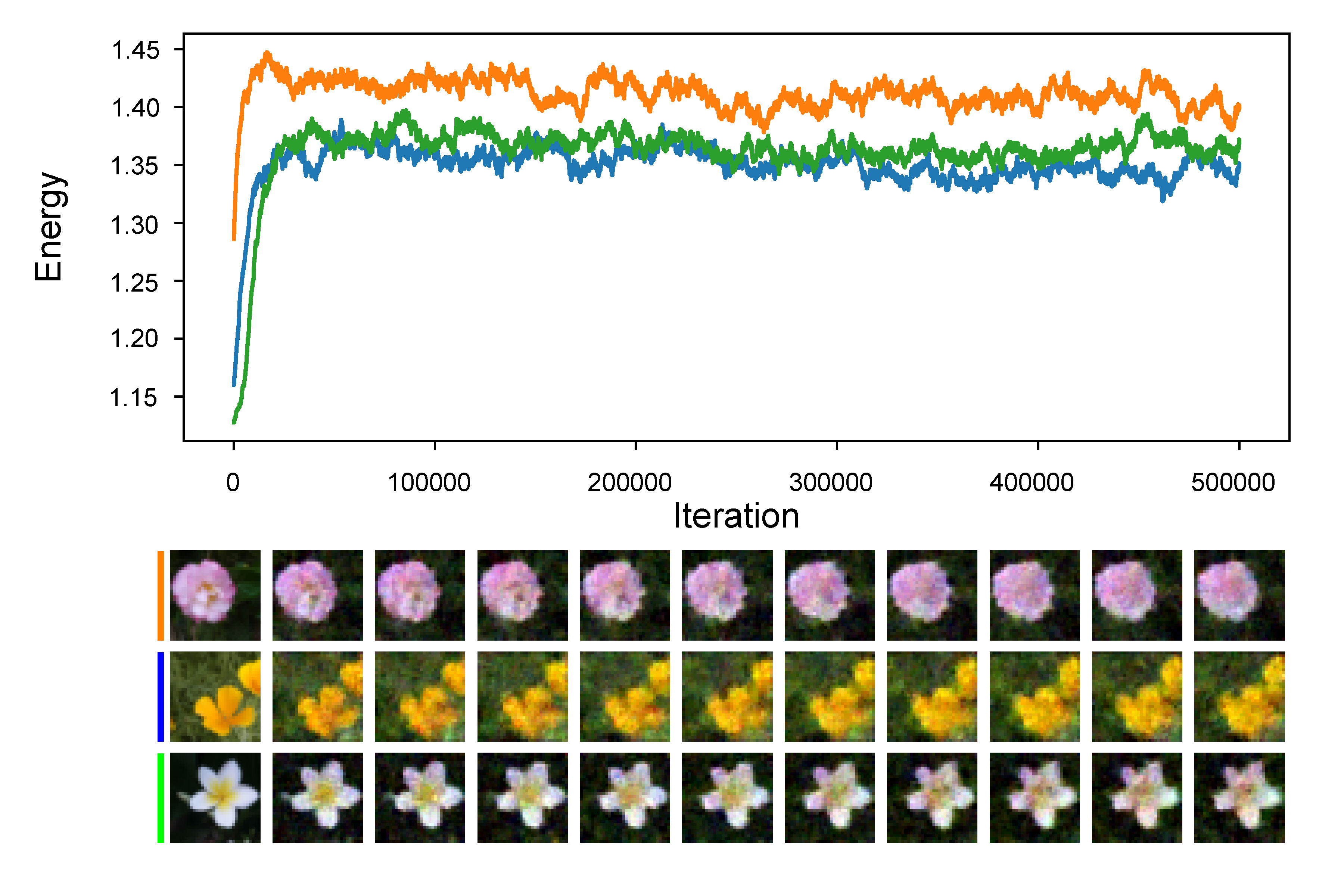}
	\caption{Long-run MH-adjusted Langevin paths from data samples to metastable samples for the Oxford Flowers 102 dataset. Models were trained with two variations of Algorithm \ref{algorithm1}: non-convergent ML trained with $L=100$ MCMC steps from noise initialization (\textit{top}), and convergent ML trained with $L=500$ MCMC steps from persistent initialization (\textit{bottom}).}
	\label{fig:energy}
\end{figure}

\begin{figure}[t]
	\centering
	\begin{tabular}{c c c}
		\small{W-GAN} & \small{WINN} & \small{Conditional EBM} \\
		\includegraphics[width=.13\textwidth]{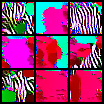} &
		\includegraphics[width=.13\textwidth]{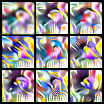} &
		\includegraphics[width=.13\textwidth]{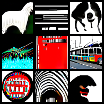}
	\end{tabular} 
	\caption{Long-run Langevin samples of recent energy-based models. Probability mass is concentrated on images that have unrealistic appearance. From left to right: Wasserstein-GAN critic on Oxford flowers \cite{arjovsky2017wasserstein}, WINN on Oxford flowers \cite{tu2018wasser}, conditional EBM on ImageNet \cite{du2019implicit}. The W-GAN critic is not trained to be an unnormalized density but we include samples for reference.}
	\label{fig:defects}
\end{figure}

However, we find that energy functions learned by prior works have a major defect regardless of MCMC initialization, network structure, and auxiliary training parameters. The long-run and steady-state MCMC samples of energy functions from all previous implementations are oversaturated images with significantly lower energy than the observed data (see Figure~\ref{fig:energy} top, and Figure~\ref{fig:defects}). In this case it is not appropriate to describe the learned model as an approximate density for the training set because the model assigns disproportionately high probability mass to images which differ dramatically from observed data. The systematic difference between high-quality short-run samples and low-quality long-run samples is a crucial phenomenon that appears to have gone unnoticed in previous studies.

\subsection{Our Contributions}

In this work, we present a fundamental understanding of learning ConvNet potentials by MCMC-based ML. We diagnose previously unrecognized complications that arise during learning and distill our insights to train models with new capabilities. Our main contributions are:
\begin{itemize}
	\item Identification of two distinct axes which characterize each parameter update in MCMC-based ML learning: 1) energy difference of positive and negative samples, and 2) MCMC convergence or non-convergence. Contrary to common expectations, convergence is \emph{not} needed for high-quality synthesis. See Figure~\ref{fig:idea} and Section~\ref{sec:two_axes}.
	\item The first ConvNet potentials trained using ML with purely noise-initialized MCMC.  Unlike prior models, our model can efficiently generate realistic and diverse samples after training from noise alone. See Figure~\ref{fig:sample}. This method is further explored in our companion work \cite{nijkamp2019learning}.
	\item The first ConvNet potentials with realistic steady-state samples. To our knowledge, ConvNet potentials with realistic MCMC sampling in the image space are unobtainable by all previous training implementations. We refer to \cite{kumar2019maximum} for a discussion. See Figure~\ref{fig:energy} (bottom) and Figure~\ref{fig:ML_valid} (middle and right column).
	\item Mapping the macroscopic structure of image space energy functions using diffusion in a magnetized energy landscape for unsupervised cluster discovery. See Figure~\ref{fig:DG}.
\end{itemize}

\subsection{Related Work}\label{sec:prior}

\subsubsection{Energy-Based Image Models}

Energy-based models define an unnormalized probability density over a state space to represent the distribution of states in a given system. The Hopfield network \cite{hopfield1982neural} adapted the Ising energy model into a model capable of representing arbitrary observed data. The RBM (Restricted Boltzmann Machine) \cite{rbm} and FRAME (Filters, Random field, And Maximum Entropy) \cite{frame,wu2000equivalence} models introduce energy functions with greater representational capacity. The RBM uses hidden units which have a joint density with the observable image pixels. The FRAME model uses convolutional filters and histogram matching to learn data features.

The pioneering work \cite{hinton2006unsupervised} studies the hierarchical energy-based model. \cite{ngiam2011learning} is an important early work proposing feedforward neural networks to model energy functions. The energy-based model in the form of (\ref{eqn:deepframe_energy}) is introduced in \cite{dai2014generative}. Deep variants of the FRAME model \cite{xie2016theory,lu2016deepframe} are the first to achieve realistic synthesis with a ConvNet potential and Langevin sampling. Similar methods are applied in \cite{du2019implicit}. The Multi-grid model \cite{gao2018learning} learns an ensemble of ConvNet potentials for images of different scales. Learning a ConvNet potential with a generator network as approximative direct sampler is explored in \cite{kim2016deep,dai2017calibrating,coopnets_pami,xie2016cooperative,han2018divergence,kumar2019maximum}.  The works \cite{TuNIPS,lazarow2017introspective,tu2018wasser} learn a ConvNet potential in a discriminative framework.

Although many of these works claim to train the energy (\ref{eqn:deepframe_energy}) to be an approximate unnormalized density for the observed images, the resulting energy functions do not have a steady-state that reflects the data (see Figure~\ref{fig:defects}). Short-run Langevin samples from informative initialization are presented as approximate steady-state samples, but further investigation shows long-run Langevin consistently disrupts the realism of short-run images. Our work is the first to address and remedy the systematic non-convergence of all prior implementations.

\subsubsection{Energy Landscape Mapping}

The full potential of the energy-based model lies in the structure of the energy landscape. Hopfield observed that the energy landscape is a model of associative memory \cite{hopfield1982neural}. Diffusion along the potential energy manifold is analogous to memory recall because the diffusion process will gradually refine a high-energy image (an incomplete or corrupted memory) until it reaches a low-energy metastable state, which corresponds to the revised memory. Techniques for mapping and visualizing the energy landscape of non-convex functions in the physical chemistry literature \cite{becker,wales2017elm} have been applied to map the latent space of Cooperative Networks \cite{hill2018telescope}. Defects in the energy function (\ref{eqn:deepframe_energy}) from previous ML implementations prevent these techniques from being applied in the image space. Our convergent ML models enable image space mapping.

\section{Learning Energy-Based Models}

In this section, we review the established principles of the MCMC-based ML learning from prior works such as \cite{hinton2002poe,frame,xie2016theory}.

\subsection{Maximum Likelihood Estimation}

An energy-based model is a Gibbs-Boltzmann density
\begin{align}
	p_\theta (x) = \frac{1}{Z(\theta)} \, \exp\{ -  U(x ; \theta) \} \label{eqn:gibbs_density}
\end{align}
over signals $x \in \mathcal{X} \subset \mathbb{R}^N$. The energy potential $U(x ; \theta)$ belongs to a parametric family $\mathcal{U} = \{ U(\cdot \; ;  \theta) : \theta \in \Theta \}$. The intractable constant $Z(\theta) = \int_\mathcal{X} \exp\{ -U(x; \theta) \} dx$ is never used explicitly because the potential $U(x; \theta)$ provides sufficient information for MCMC sampling. In this paper we focus our attention on energy potentials with the form
\begin{align}
	U(x ; \theta) = F(x  ; \theta)  \label{eqn:deepframe_energy}
\end{align}
where $F(x ; \theta)$ is a convolutional neural network with a single output channel and weights $\theta\in \mathbb{R}^D$.

In ML learning, we seek to find $\theta \in \Theta$ such that the parametric model $p_\theta(x)$ is a close approximation of the data distribution $q(x)$. One measure of closeness is the Kullback-Leibler (KL) divergence. Learning proceeds by solving
\begin{align}\label{eq:mle}
\argmin_\theta\mathcal{L}(\theta)
&=\argmin_\theta D_{KL}(q\|p_\theta)\\
&= \argmin_\theta \left\{ \log Z(\theta) + E_q [ U(X; \theta)]\right\}. \label{eqn:ml_objective}
\end{align}
We can minimize $\mathcal{L} (\theta)$ by finding the roots of the derivative
\begin{align}
\frac{d}{d\theta} \mathcal{L} (\theta) &= \frac{d}{d \theta} \log Z(\theta) + \frac{d}{d\theta} E_q [U(X; \theta)] .
\end{align}
The term $\frac{d}{d \theta} \log Z(\theta)$ is intractable, but it can be expressed
\begin{align}
\frac{d}{d \theta} \log Z(\theta) = - E_{p_\theta} \left[ \frac{\partial}{\partial \theta} U(X; \theta)\right] .
\end{align}
The gradient used to learn $\theta$ then becomes
\begin{align}
\frac{d}{d\theta} \mathcal{L}(\theta) &=  \frac{d}{d \theta} E_q [ U(X;\theta) ] - E_{p_\theta} \left[\frac{\partial}{\partial \theta} U(X;\theta) \right] \label{eqn:ml_grad} \\
&\approx \frac{\partial}{\partial \theta} \left( \frac{1}{n} \sum_{i=1}^n U(X^+_i ; \theta) - \frac{1}{m}\sum_{i=1}^m U(X_i^-; \theta) \right) \label{eqn:ml_grad_approx}
\end{align}
where $\{X^+_i\}_{i=1}^n$ are i.i.d. samples from the data distribution $q$ (called \emph{positive} samples since probability is increased), and $\{X_i^- \}_{i=1}^m $ are i.i.d. samples from current learned distribution $p_\theta$ (called \emph{negative} samples since probability is decreased). In practice, the positive samples $\{X^+_i\}_{i=1}^n$ are a batch of training images and the negative samples $\{X_i^- \}_{i=1}^m $ are obtained after $L$ iterations of MCMC sampling.

\subsection{MCMC Sampling with Langevin Dynamics}\label{subsec:langevin}
Obtaining the negative samples $\{X_i^- \}_{i=1}^m$ from the current distribution $p_\theta$ is a computationally intensive task which must be performed for each update of $\theta$. ML learning does not impose a specific MCMC algorithm. Early energy-based models such as the RBM and FRAME model use Gibbs sampling as the MCMC method. Gibbs sampling updates each dimension (one pixel of the image) sequentially. This is computationally infeasible when training an energy with the form (\ref{eqn:deepframe_energy}) for standard image sizes.

Several works studying the energy (\ref{eqn:deepframe_energy}) recruit Langevin Dynamics to obtain the negative samples \cite{xie2016theory,lu2016deepframe,xie2016cooperative,gao2018learning,tu2018wasser}. The Langevin Equation
\begin{align}
X_{\ell+1} = X_\ell - \frac{\varepsilon^2}{2} \frac{\partial}{\partial x} \, U(X_\ell ; \theta) + \varepsilon Z_\ell , \label{eqn:langevin}
\end{align}
where $Z_\ell \sim \textrm{N}(0, \, I_N )$ and $\varepsilon>0$, has stationary distribution $p_\theta$ \cite{geman84gibbs,neal2011mcmc}. A complete implementation of Langevin Dynamics requires a momentum update and Metropolis-Hastings update in addition to (\ref{eqn:langevin}), but most authors find that these can be ignored in practice for small enough $\varepsilon$ \cite{chen104hmc}.

Like most MCMC methods, Langevin dynamics exhibits high auto-correlation and has difficulty mixing between separate modes. Even so, long-run Langevin samples with a suitable initialization can still be considered approximate steady-state samples, as discussed next.

\subsection{MCMC Initialization}\label{subsec:mcmc_priming}
We distinguish two main branches of MCMC initialization: \emph{informative initialization}, where the density of initial states is meant to approximate the model density, and \emph{non-informative initialization}, where initial states are obtained from a distribution that is unrelated to the model density. \emph{Noise initialization} is a specific type of non-informative initialization where initial states come from a noise distribution such as uniform or Gaussian.

In the most extreme case, a Markov chain initialized from its steady-state will follow the steady-state distribution after a single MCMC update. In more general cases, a Markov chain initialized from an image that is likely under the steady-state can converge much more quickly than a Markov chain initialized from noise. For this reason, all prior works studying ConvNet potentials use informative initialization.

\emph{Data-based initialization} uses samples from the training data as the initial MCMC states. Contrastive Divergence (CD) \cite{hinton2002poe} introduces this practice. The Multigrid Model \cite{gao2018learning} generalizes CD by using multi-scale energy functions to sequentially refine downsampled data.

\emph{Persistent initialization} uses negative samples from a previous learning iteration as initial MCMC states in the current iteration. The persistent chains can be initialized from noise as in \cite{frame,lu2016deepframe,xie2016theory} or from data samples as in Persistent Contrastive Divergence (PCD) \cite{pcd}. The Cooperative Learning model \cite{xie2016cooperative} generalizes persistent chains by learning a generator for proposals in tandem with the energy.

In this paper we consider long-run Langevin chains from both data-based initialization such as CD and persistent initialization such as PCD to be approximate steady-state samples, even when Langevin chains cannot mix between modes. Prior art indicates that both initialization types span the modes of the learned density, and long-run Langevin samples will travel in a way that respects the $p_{\theta}$ in the local landscape.

Informative MCMC initialization during ML training can limit the ability of the final model $p_\theta$ to generate new and diverse synthesized images after training. MCMC samples initialized from noise distributions after training tend to result in images with a similar type of appearance when informative initialization is used in training. 

In contrast to common wisdom, we find that informative initialization is not necessary for efficient and realistic synthesis when training ConvNet potentials with ML. In accordance with common wisdom, we find that informative initialization is essential for learning a realistic steady-state.

\section{Two Axes of ML Learning}\label{sec:two_axes}

Inspection of the gradient (\ref{eqn:ml_grad_approx}) reveals the central role of the difference of the average energy of negative and positive samples. Let
\begin{align}
d_{s_t} (\theta) &= E_q [ U(X; \theta) ] - E_{s_t} [U(X; \theta) ] \label{eqn:ml_d_function}
\end{align} 
where $s_t (x)$ is the distribution of negative samples given the finite-step MCMC sampler and initialization used at training step $t$.  The difference $d_{s_t} (\theta)$ measures whether the positive samples from the data distribution $q$ or the negative samples from $s_t$ are more likely under the model $p_\theta$. The ideal case $p_\theta = q$ (perfect learning) and $s_t = p_\theta$ (exact MCMC convergence) satisfies $d_{s_t} (\theta)= 0$. A large value of $| d_{s_t} |$ indicates that either learning or sampling (or both) have not converged.

Although $d_{s_t} (\theta)$ is not equivalent to the ML objective (\ref{eqn:ml_objective}), it bridges the gap between theoretical ML and the behavior encountered when MCMC approximation is used. Two outcomes occur for each update on the parameter path $\{ \theta_t \}_{t=1}^{T+1}$:
\begin{enumerate}
	\item $d_{s_t} (\theta_{t}) < 0 $ (expansion) or $d_{s_t} (\theta_{t}) > 0$ (contraction)
	\item $s_t \approx p_{\theta_t}$ (MCMC convergence) or $s_t \not\approx p_{\theta_t}$ (MCMC non-convergence) .
\end{enumerate}

We find that only the first axis governs the stability and synthesis results of the learning process. Oscillation of expansion and contraction updates is an indicator of stable ML learning, but this can occur in cases where either $s_t$ is always approximately convergent or where $s_t$ never converges.

Behavior along the second axis determines the realism of steady-state samples from the final learned energy.  Samples from $p_{\theta_t}$ will be realistic if and only if $s_t$ has realistic samples and $s_t \approx p_{\theta_t}$. We use \emph{convergent ML} to refer to implementations where $s_t \approx p_{\theta_t}$ for all $t> t_0$, where $t_0$ represents burn-in learning steps (e.g. early stages of persistent learning). We use \emph{non-convergent ML} to refer to all other implementations. All prior ConvNet potentials are learned with non-convergent ML, although this is not recognized by previous authors.

Without proper tuning of the sampling phase, the learning heavily gravitates towards non-convergent ML. In this section we outline principles to explain this behavior and provide a remedy for the tendency of model non-convergence.

\subsection{First Axis: Expansion or Contraction}
Following prior art for high-dimensional image models, we use the Langevin Equation (\ref{eqn:langevin}) to obtain MCMC samples. Let $w_t$ give the joint distribution of a Langevin chain $(Y_t^{(0)}, \dots, Y_t^{(L)})$ at training step $t$, where $Y_t^{(\ell+1)}$ is obtained by applying (\ref{eqn:langevin}) to $Y_t^{(\ell)}$ and $Y_t^{(L)} \sim s_t$. Since the gradient $\frac{\partial U}{\partial x}$ appears directly in the Langevin equation, the quantity 
\[
v_t = E_{w_t}\left[ \frac{1}{L+1} \sum_{\ell=0}^{L} \left\| \frac{\partial}{\partial y} U(Y_t^{(\ell)}; \theta_t )\right\|_2 \right] , 
\] 
which gives the average image gradient magnitude of $U$ along an MCMC path at training step $t$, plays a central role in sampling. Sampling at noise magnitude $\varepsilon$ will lead to very different behavior depending on the gradient magnitude. If $v_t$ is very large, gradients will overwhelm the noise and the resulting dynamics are similar to gradient descent. If $v_t$ is very small, sampling becomes an isotropic random walk. A valid image density should appropriately balance energy gradient magnitude and noise strength to enable realistic long-run sampling.

\begin{figure}[ht!]
	\centering
	\includegraphics[width=.46\textwidth]{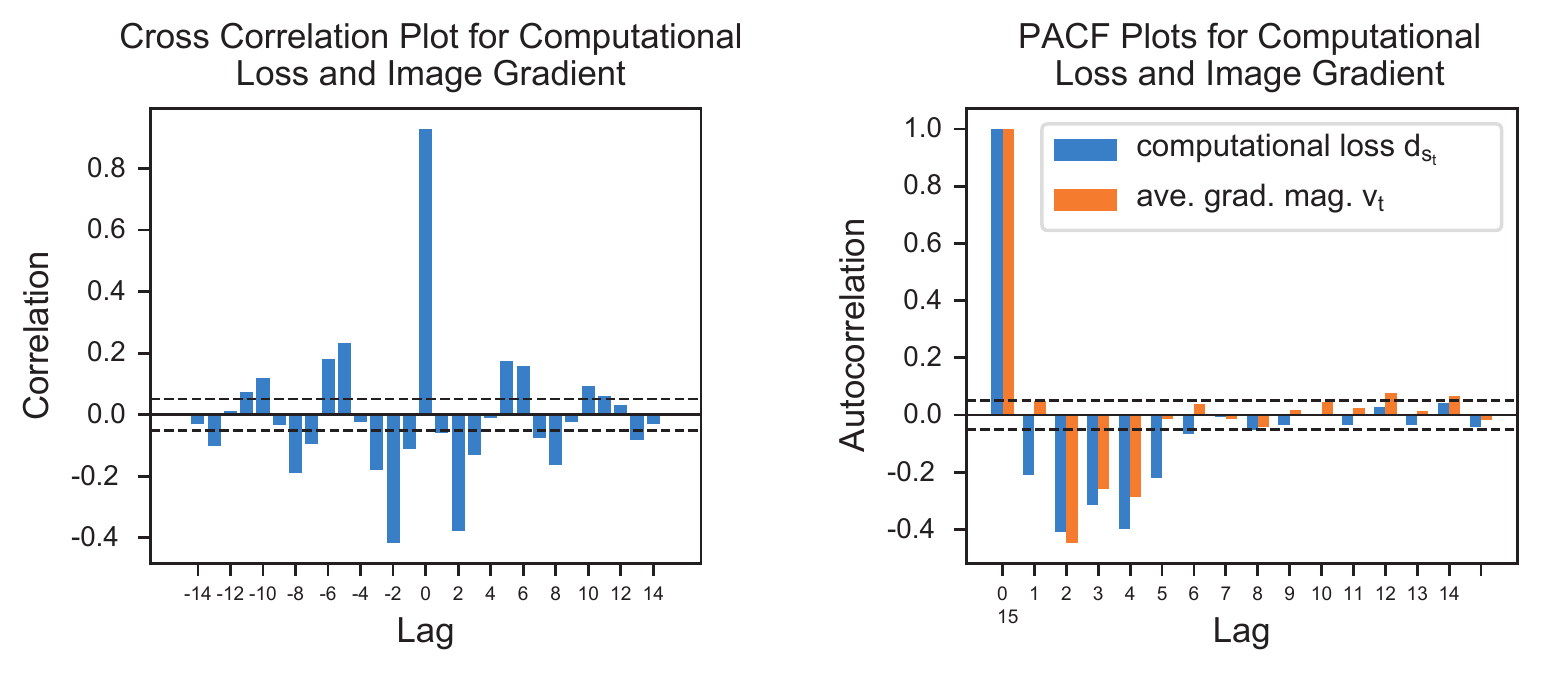}
	\caption{Illustration of expansion/contraction oscillation for a single training implementation. This behavior is typical of convergent \emph{and} non-convergent ML. \emph{Left:} Cross correlation of $d_{s_t}$ (uncentered) and $v_t$ (mean centered). The two are highly correlated at lag 0 and exhibit negative correlation for lag $\pm 3$ steps, indicating that expansion updates tend to increase gradient strength in the near future and vice-versa.  \emph{Right:} PACF plots of $d_{s_t}$ (uncentered) and $v_t$ (mean centered). Both have a strong negative autocorrelation within the next 4 training batches, showing that expansion updates tend to follow contraction updates and vice-versa.}
	\label{fig:spectral_diagnostic}
\end{figure}

We empirically observe that expansion and contraction updates tend to have opposite effects on $v_t$ (see Figure \ref{fig:spectral_diagnostic}). Gradient magnitude $v_t$ and computational loss $d_{s_t}$ are highly correlated at the current iteration and exhibit significant negative correlation at a short-range lag. Both have significant negative autocorrelation for short-range lag. This indicates that expansion updates tend to increase $v_t$ and contraction updates tend to decrease $v_t$, and that expansion updates tend to lead to contraction updates and vice-versa. We believe that the natural oscillation between expansion and contraction updates underlies the stability of ML with (\ref{eqn:deepframe_energy}). 

Learning can become unstable when $U$ is updated in the expansion phase for many consecutive iterations if $v_t \rightarrow \infty$ and as $U(X^+) \rightarrow -\infty$ for positive samples and $U(X^- ) \rightarrow \infty$ for negative samples. This behavior is typical of W-GAN training (interpreting the generator as $w_t$ with $L=0$) and the W-GAN Lipschitz bound is needed to prevent such instability. In ML learning with ConvNet potentials, consecutive updates in the expansion phase will increase $v_t$ so that the gradient can better overcome noise and samples can more quickly reach low-energy regions. In contrast, many consecutive contraction updates can cause $v_t$ to shrink to 0, leading to the solution $U(x) = c$ for some constant $c$ (see Figure \ref{fig:convergent_diagnostic} right, blue lines). In proper ML learning, the expansion updates that follow contraction updates prevent the model from collapsing to a flat solution and force $U$ to learn meaningful features of the data.

Throughout our experiments, we find that the network can easily learn to balance the energy of the positive and negative samples so that $d_{s_t} (\theta_t) \approx 0$ after only a few model updates. In fact, ML learning can easily adjust $v_t$ so that the gradient is strong enough to balance $d_{s_t}$ and obtain high-quality samples from virtually \emph{any} initial distribution in a small number of MCMC steps. This insight leads to our ML method with noise-initialized MCMC. The natural oscillation of ML learning is the foundation of the robust synthesis capabilities of ConvNet potentials, but realistic short-run MCMC samples can mask the true steady-state behavior.

\subsection{Second Axis: MCMC Convergence or Non-Convergence}

In the literature, it is expected that the finite-step MCMC distribution $s_t$ must approximately converge to its steady-state $p_{\theta_t}$ for learning to be effective. On the contrary, we find that high-quality synthesis is possible, and actually easier to learn, when there is a drastic difference between the finite-step MCMC distribution $s_t$ and true steady-state samples of $p_{\theta_t}$. An examination of ConvNet potentials learned by existing methods shows that in all cases, running the MCMC sampler for significantly longer than the number of training steps results in samples with significantly lower energy and unrealistic appearance. Although synthesis is possible without convergence, it is not appropriate to describe a non-convergent ML model $p_\theta$ as an approximate data density.

\begin{figure}
	\centering
	\includegraphics[width=.45\textwidth]{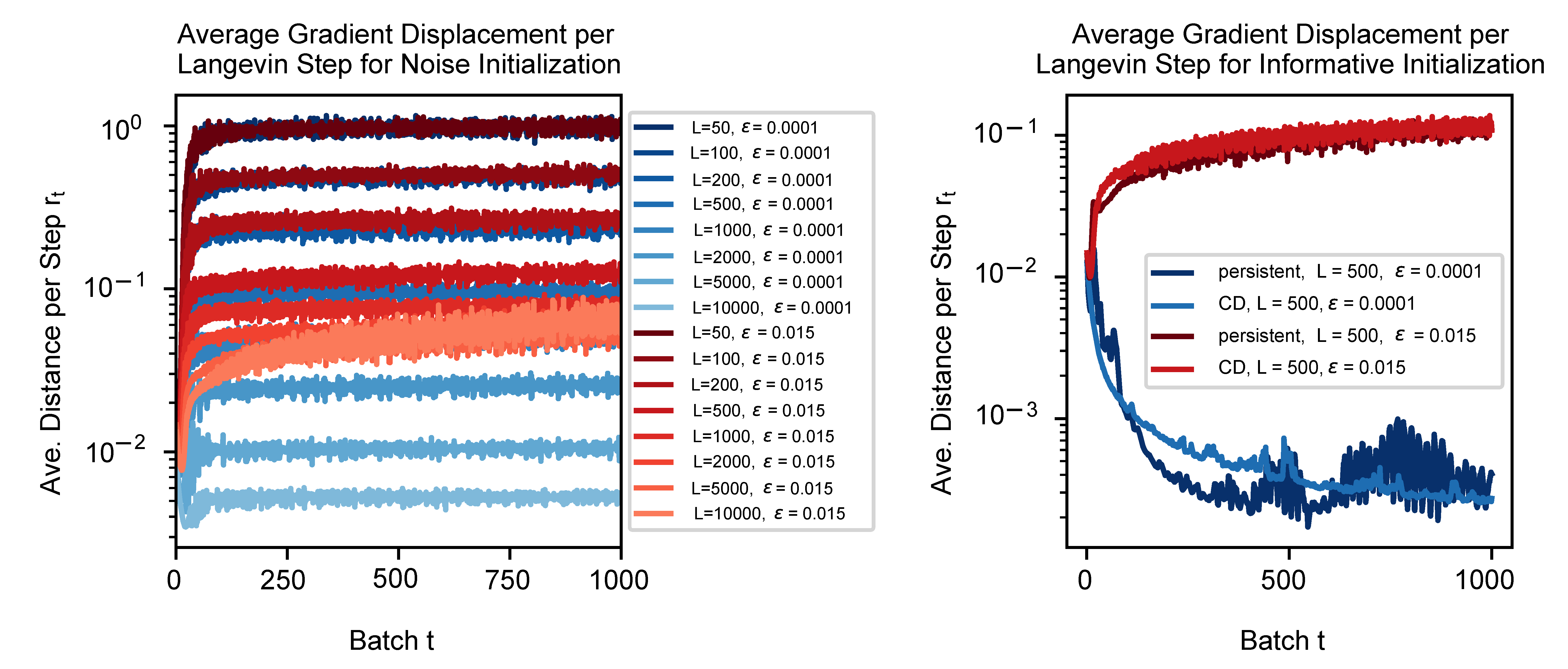}
	\caption{Illustration of gradient strength for convergent and non-convergent ML. With low noise (blue) the energy either learns only the burn-in path (left) or contracts to a constant function (right). With sufficient noise (red), the network gradient learns to balance with noise magnitude and it becomes possible to learn a realistic steady-state.}
	\label{fig:convergent_diagnostic}
\end{figure}

Oscillation of expansion and contraction updates occurs for both convergent and non-convergent ML learning, but for very different reasons. In convergent ML, we expect the average gradient magnitude $v_t$  to converge to a constant that is balanced with the noise magnitude $\varepsilon$ at a value that reflects the temperature of the data density $q$. However, ConvNet potentials can circumvent this desired behavior by tuning $v_t$ with respect to the burn-in energy landscape rather than noise $\varepsilon$. Figure \ref{fig:convergent_diagnostic} shows how average image space displacement $r_t = \frac{\varepsilon^2}{2} v_t$ is affected by noise magnitude $\varepsilon$ and number of Langevin steps $L$ for noise, data-based, and persistent MCMC initializations. 

For noise initialization with low $\varepsilon$, the model adjusts $v_t$ so that $r_t L \approx R$ where $R$ is the average distance between an image from the noise initialization distribution and an image from the data distribution. In other words, the MCMC paths obtained from non-convergent ML with noise initialization are nearly linear from the starting point to the ending point. Mixing does \emph{not} improve when $L$ increases because $r_t$ shrinks in proportion to the increase. Oscillation of expansion and contraction updates occurs because the model tunes $v_t$ to control how far along the burn-in path the negative samples travel. Samples never reach the steady-state energy spectrum and MCMC mixing is not possible.

For data initialization and persistent initialization with low $\varepsilon$, we see that $v_t, r_t \rightarrow 0$ and that learning tends to the trivial solution $U (x) = c$. This occurs because contraction updates dominate the learning dynamics. At low $\varepsilon$, samples initialized from the data will easily have lower energy than the data since sampling reduces to gradient descent.  To our knowledge no authors have trained (\ref{eqn:deepframe_energy}) using CD, possibly because the energy can easily collapse to a trivial flat solution. For persistent learning, the model learns to synthesize meaningful features early in learning and then contracts in gradient strength once it becomes easy to find negative samples with lower energy than the data. Previous authors who trained models with persistent chains use auxiliary techniques such as a Gaussian prior \cite{xie2016theory} or occasional rejuvenation of chains from noise \cite{du2019implicit} which prevent unbalanced network contraction, although the role of these techniques is not recognized by the authors.

For all three initialization types, we can see that convergent ML becomes possible when $\varepsilon$ is large enough. ML with noise initialization behaves similarly for high and low $\varepsilon$ when $L$ is small. For large $L$ with high $\varepsilon$, the model tunes $v_t$ to balance with $\varepsilon$ rather than $R / L$. The MCMC samples complete burn-in and begin to mix for large $L$, and increasing $L$ will indeed lead to improved MCMC convergence as usual. For data-based and persistent initialization, we see that $v_t$ adjusts to balance with $\varepsilon$ instead of contracting to 0 because the noise added during Langevin sampling forces $U$ to learn meaningful features. 

\begin{figure}[h]
	\centering
	\hspace*{-.25cm}\includegraphics[width=.48\textwidth]{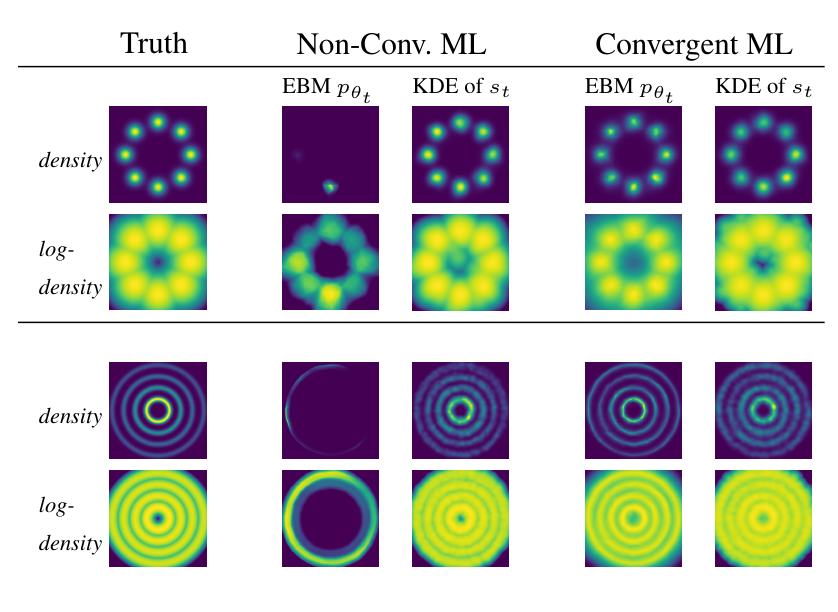}

	\caption{Comparison of convergent and non-convergent ML for 2D toy distributions. Non-convergent ML does not learn a valid density but the kernel density estimate of the negative samples reflects the groundtruth. Convergent ML learns an energy that closely approximates the true density.}
	\label{fig:toy_exp}
\end{figure}

\subsection{Learning Algorithm}

We now present an algorithm for ML learning. The algorithm is essentially the same as earlier works such as \cite{xie2016theory} that investigate the potential (\ref{eqn:deepframe_energy}). Our intention is not to introduce a novel algorithm but to demonstrate the range of phenomena that can occur with the ML objective based on changes to MCMC sampling. We present guidelines for the effect of tuning on the learning outcome.

\begin{algorithm}
		\SetKwInOut{Input}{input} \SetKwInOut{Output}{output}
		\DontPrintSemicolon
		\Input{ConvNet potential $U(x; \theta)$, number of training steps $T$, initial weight $\theta_1$, training images $\{x^+_i \}_{i=1}^{N_\textrm{data}}$,  step size $\varepsilon$, noise indicator $\tau \in \{ 0, 1 \}$, Langevin steps $L$, learning rate $\gamma$.}
		\Output{Weights $\theta_{T+1}$ for energy $U(x ; \theta)$.}
		\For{$t = 1:{T}$}{
						
			\smallskip
			1. Draw batch images $\{ X_i^+ \}_{i=1}^n$ from training set. Draw initial negative samples $\{ Y_i^{(0)} \}_{i=1}^m$ from MCMC initialization method (noise or informative initialization, see Section~\ref{subsec:mcmc_priming}). \;
			2. Update $\{ Y_i^{(0)} \}_{i=1}^m$ with
			\begin{align*}
			Y^{(\ell)}_i = Y^{(\ell-1)}_i - \frac{\varepsilon^2}{2} \frac{\partial}{\partial y} U(Y^{(\ell-1)}_i ; \theta_{t}) + \varepsilon \tau Z_{i, \ell} ,
			\end{align*}
			 where $Z_{i , \ell} \sim \textrm{N}(0, I_N)$, for $L$ steps to obtain negative samples $\{ X_i^- \}_{i=1}^m = \{ Y_i^{(L)} \}_{i=1}^m$. \;
			3.  Update the weights by $\theta_{t+1} = \theta_{t} - g(\Delta \theta_t, \gamma)$ where $\Delta \theta_t $ is the stochastic gradient (\ref{eqn:ml_grad_approx}) and $g$ is the SGD or Adam \cite{kingma2015adam} optimizer.
		}
		\caption{ML Learning}
		\label{algorithm1}
\end{algorithm}

\begin{figure*}
	\centering
	\includegraphics[width=.85\textwidth]{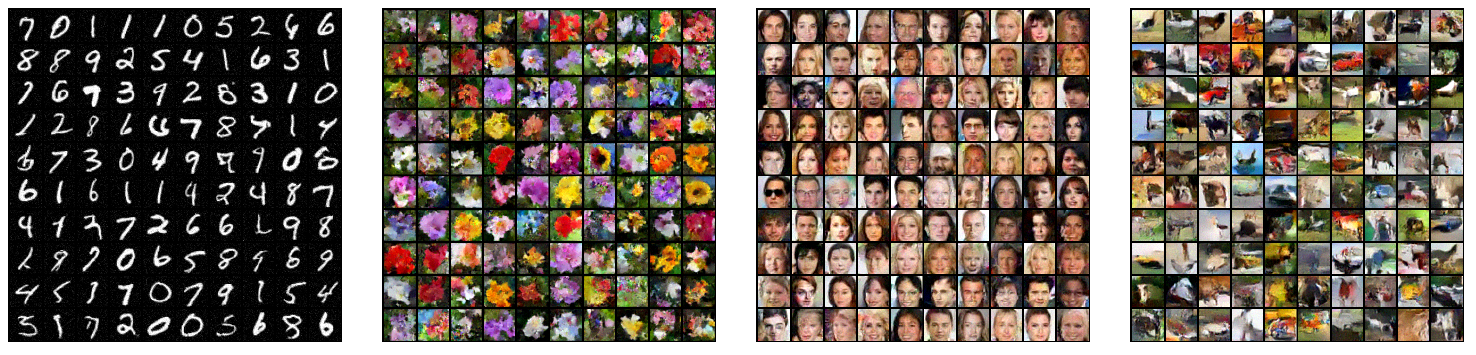}	
	\caption{Short-run samples obtained from an energy function trained with non-convergent ML with noise initialization. The images are generated using 100 Langevin updates from uniform noise initialization. Contrary to prior art, informative initialization is not needed for high-quality synthesis. From left to right: MNIST, Oxford Flowers 102, CelebA, CIFAR-10.}
	\label{fig:sample}
\end{figure*}

\begin{itemize}
	\item \textit{Noise and Step Size for Non-Convergent ML}: For non-convergent training we find the tuning of noise and step-size have little effect on training stability. We use $\varepsilon = 1$ and $\tau = 0$. Noise is not needed for oscillation because $d_{s_t}$ is controlled by the depth of samples along the burn-in path. Including low noise appears to improve synthesis quality.
	\item \textit{Noise and Step Size for Convergent ML:} For convergent training, we find that it is essential to include noise with $\tau=1$ and precisely tune $\varepsilon$ so that the network learns true mixing dynamics through the gradient strength. The step size $\varepsilon$ should approximately match the local standard deviation of the data along the most constrained direction \cite{neal2011mcmc}. An effective $\varepsilon$ for $32\times32$ images with pixel values in [-1, 1] appears to lie around $0.015$. 
	\item \textit{Number of Steps}: When $\tau = 0$ or $\tau = 1$ and $\varepsilon$ is very small, learning leads to similar non-convergent ML outcomes for any $L \ge 100$. When $\tau=1$ and $\varepsilon$ is correctly tuned, sufficiently high values of $L$ lead to convergent ML and lower values of $L$ lead to non-convergent ML.
	\item \textit{Informative Initialization:} Informative MCMC initialization is not needed for non-convergent ML even with as few as $L=100$ Langevin updates. The model can naturally learn fast pathways to realistic negative samples from an arbitrary initial distribution. On the other hand, informative initialization can greatly reduce the magnitude of $L$ needed for convergent ML. We use persistent initialization starting from noise.
	\item \textit{Network structure}: For the first convolutional layer, we observe that a $3\times 3$ convolution with stride $1$ helps to avoid checkerboard patterns or other artifacts. For convergent ML, use of non-local layers \cite{wang2018nonlocal} appears to improve synthesis realism.
	\item \textit{Regularization and Normalization}: Previous studies employ a variety of auxiliary training techniques such as prior distributions (e.g. Gaussian), weight regularization, batch normalization, layer normalization, and spectral normalization to stabilize sampling and weight updates. We find that these techniques are not needed.
	\item \textit{Optimizer and Learning Rate:} For non-convergent ML, Adam improves training speed and image quality. Our non-convergent models use Adam with $\gamma = 0.0001$. For convergent ML, Adam appears to interfere with learning a realistic steady-state and we use SGD instead. When using SGD with $\tau=1$ and properly tuned $\varepsilon$ and $L$, higher values of $\gamma$ lead to non-convergent ML and sufficiently low values of $\gamma$ lead to convergent ML.
\end{itemize}

\section{Experiments}

\subsection{Low-Dimensional Toy Experiments}
We first demonstrate the outcomes of convergent and non-convergent ML for low-dimensional toy distributions (Figure~\ref{fig:toy_exp}). Both toy models have a standard deviation of $0.15$ along the most constrained direction, and the ideal step size for Langevin dynamics is close to this value \cite{neal2011mcmc}. Non-convergent models are trained using noise MCMC initialization with $L=100$ and $\varepsilon = 0.01$ (too low for the data temperature) and convergent models are trained using persistent MCMC initialization with $L=500$ and $\varepsilon = 0.125$ (approximately the right magnitude relative to the data temperature). The distributions of the short-run samples from the non-convergent models reflect the ground-truth densities, but the learned densities are sharply concentrated and different from the ground-truths. In higher dimensions this sharp concentration of non-convergent densities manifests as oversaturated long-run images. With sufficient Langevin noise, one can learn an energy function that closely approximates the ground-truth.

\subsection{Synthesis from Noise with Non-Convergent ML Learning}
In this experiment, we learn an energy function (\ref{eqn:deepframe_energy}) using ML with uniform noise initialization and short-run MCMC. We apply our ML algorithm with $L =100$ Langevin steps starting from uniform noise images for each update of $\theta$ with $\tau=0$ and $\varepsilon=1$. We use Adam with $\gamma=0.0001$.

\begin{figure}[h]
	\centering
	\includegraphics[width=.45\textwidth]{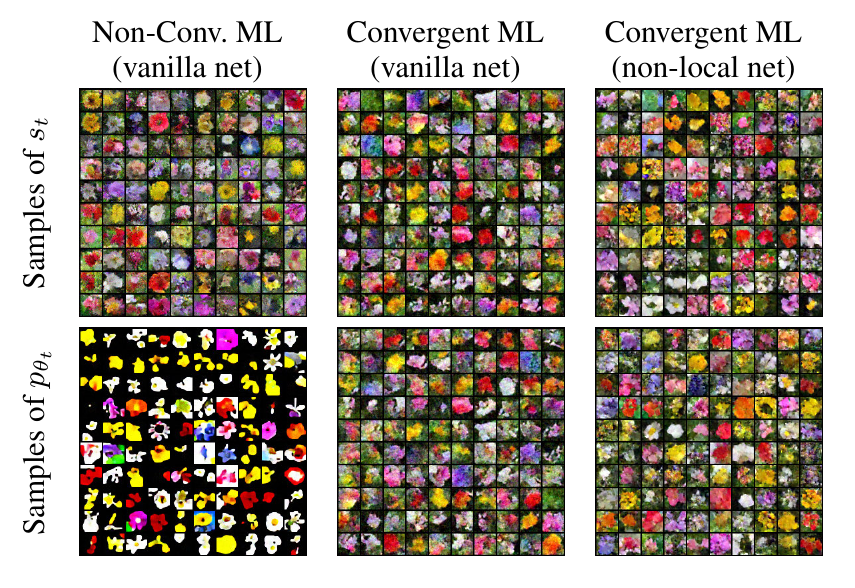}
	\caption{Comparison of negative samples and steady-state samples. Method: non-convergent ML using noise initialization and 100 Langevin steps (\emph{left}), convergent ML with a vanilla ConvNet, persistent initialization and 500 Langevin steps (\emph{center}), and convergent ML with a non-local net, persistent initialization and 500 Langevin steps (\emph{right}).}
	\label{fig:ML_valid}
\end{figure}

Previous authors argued that informative MCMC initialization is a key element for successful synthesis with ML learning, but our learning method can sample from scratch with the same Langevin budget. Unlike the models learned by previous authors, our models can generate high-fidelity and diverse images from a noise signal. Our results are shown in Figure~\ref{fig:sample}, Figure~\ref{fig:ML_valid} (left), and Figure~\ref{fig:energy} (top). Our recent companion work \cite{nijkamp2019learning} thoroughly explores the capabilities of noise-initialized non-convergent ML.

\subsection{Convergent ML Learning}

With the correct Langevin noise, one can ensure that MCMC samples mix in the steady-state energy spectrum throughout training. The model will eventually learn a realistic steady-state as long as MCMC samples approximately converge for each parameter update $t$ beyond a burn-in period $t_0$. One can implement convergent ML with noise initialization, but we find that this requires $L \approx$ 20,000 steps.

Informative initialization can dramatically reduce the number of MCMC steps needed for convergent learning. By using SGD with learning rate $\gamma = 0.0005$, noise indicator $\tau = 1$ and step size $\varepsilon = 0.015$, we were able to train convergent models using persistent initialization and $L = 500$ sampling steps. We initialize 10,000 persistent images from noise and update 100 images for each batch. We implement the same training procedure for a vanilla ConvNet and a network with non-local layers \cite{wang2018nonlocal}. Our results are shown in Figure~\ref{fig:ML_valid} (middle, right) and Figure~\ref{fig:energy} (bottom).

\begin{figure}[h]
	\centering
	\includegraphics[width=.43\textwidth]{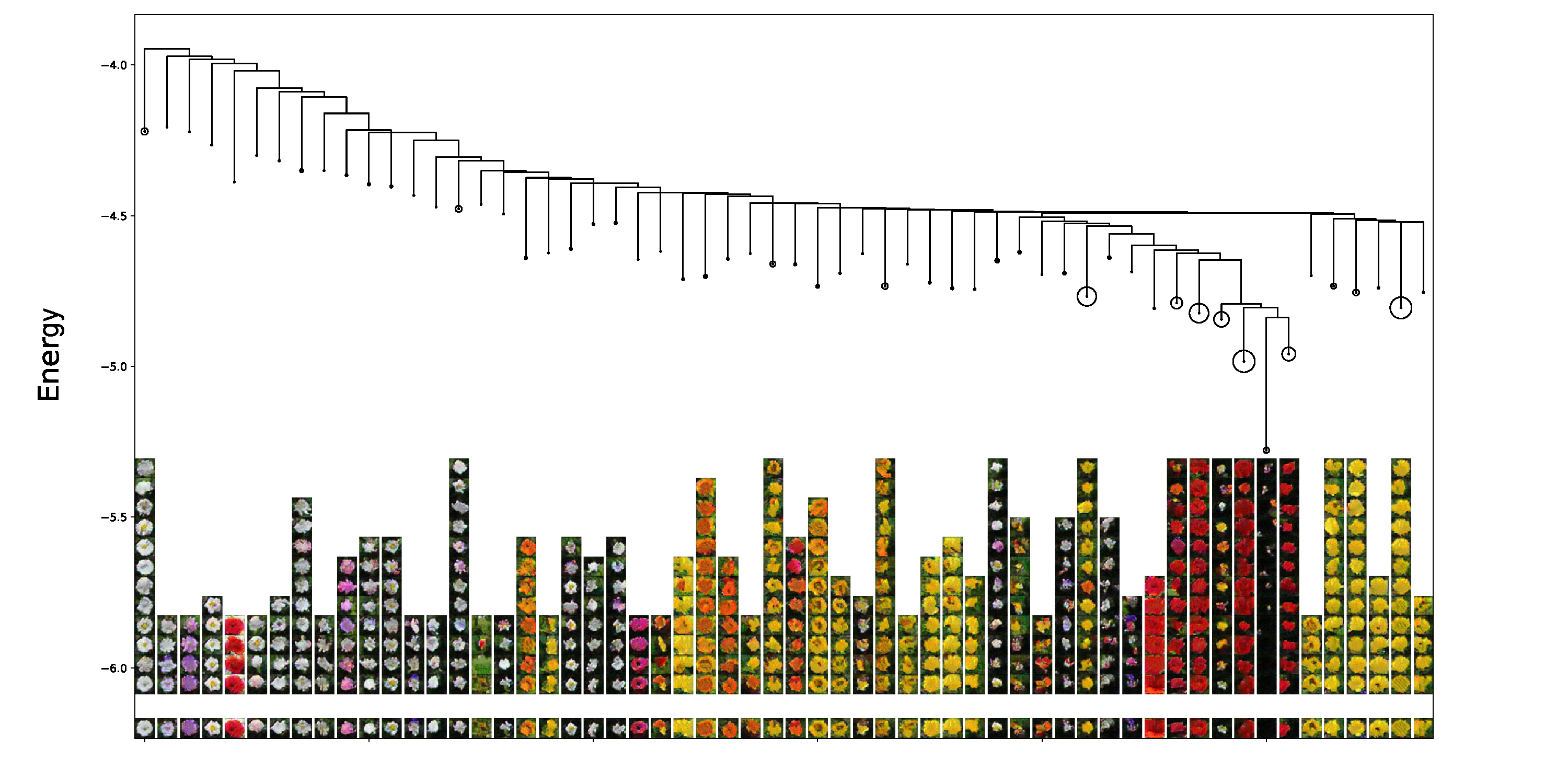}
	\caption{Visualization of basin structure of the learned energy function $U(x)$ for the Oxford Flowers 102 dataset. Columns display randomly selected basins members and circles indicate the total number of basin members. Vertical lines encode basin minimum energy and horizontal lines depict the lowest known barrier at which two basins merge.}
	\label{fig:DG}
\end{figure}

\subsection{Mapping the Image Space}

A well-formed energy function partitions the image space into meaningful Hopfield basins of attraction. Following Algorithm 3 of \cite{hill2018telescope}, we map the structure of a convergent energy. We first identify many metastable MCMC samples. We then sort the metastable samples from lowest energy to highest energy and sequentially group images if travel between samples is possible in a magnetized energy landscape. This process is continued until all minima have been clustered. Our mappings show that the convergent energy has meaningful metastable structures encoding recognizable concepts (Figure \ref{fig:DG}).

\section{Conclusion and Future Work}
Our experiments on energy-based models with the form (\ref{eqn:deepframe_energy}) reveal two distinct axes of ML learning. We use our insights to train models with sampling capabilities that are unobtainable by previous implementations. The informative MCMC initializations used by previous authors are not necessary for high-quality synthesis. By removing this technique we train the first energy functions capable of high-diversity and realistic synthesis from noise initialization after training. We identify a severe defect in the steady-state distributions of prior implementations and introduce the first ConvNet potentials of the form (\ref{eqn:deepframe_energy}) for which steady-state samples have realistic appearance. Our observations could be very useful for convergent ML learning with more complex MCMC initialization methods used in \cite{xie2016cooperative,gao2018learning}. We hope that our work paves the way for future unsupervised and weakly supervised applications with energy-based models.

\section*{Acknowledgment}
The work is supported by DARPA XAI project N66001-17-2-4029; ARO project W911NF1810296; and ONR MURI project N00014-16-1-2007; and Extreme Science and Engineering Discovery Environment (XSEDE) grant ASC170063. We thank Prafulla Dhariwal and Anirudh Goyal for helpful discussions.

\bibliography{main}
\bibliographystyle{aaai}
\end{document}